\documentclass[shortpaper]{clv3}

\usepackage{hyperref}
\usepackage{xcolor}
\definecolor{darkblue}{rgb}{0, 0, 0.5}
\hypersetup{colorlinks=true,citecolor=darkblue, linkcolor=darkblue, urlcolor=darkblue}

\usepackage{latexsym}
\usepackage{wrapfig}
\usepackage{amsmath}
\usepackage{float}
\usepackage{dblfloatfix}
\usepackage{subfigure}
\usepackage{url}
\usepackage{graphicx}
\usepackage{amsfonts}
\usepackage{booktabs}
\usepackage{multirow}
\usepackage{multicol}
\usepackage{dashrule}
\usepackage{adjustbox}
\usepackage{tabularx} 
\usepackage{colortbl}

\usepackage{CJKutf8}

\usepackage{cleveref} 

\crefformat{section}{\S#2#1#3}
\crefformat{subsection}{\S#2#1#3}
\crefformat{subsubsection}{\S#2#1#3}
\crefrangeformat{section}{\S\S#3#1#4 to~#5#2#6}
\crefmultiformat{section}{\S\S#2#1#3}{ and~#2#1#3}{, #2#1#3}{ and~#2#1#3}

\issue{1}{1}{2016}

\dochead{ }

\runningtitle{Incorporating Relevant Knowledge in Context Modeling and Response Generation}

\runningauthor{ }

\begin{document}

\title{Incorporating Relevant Knowledge in Context Modeling and Response Generation}

\author{Yanran Li}
\affil{The Hong Kong Polytechnic University}

\author{Wenjie Li}
\affil{The Hong Kong Polytechnic University}

\author{Ziqiang Cao}
\affil{The Hong Kong Polytechnic University}

\author{Chengyao Chen}
\affil{The Hong Kong Polytechnic University}

\maketitle

\begin{abstract}
To sustain engaging conversation, it is critical for chatbots to make good use of relevant knowledge. Equipped with a knowledge base, chatbots are able to extract conversation-related attributes and entities to facilitate context modeling and response generation. In this work, we distinguish the uses of attribute and entity and incorporate them into the encoder-decoder architecture in different manners. Based on the augmented architecture, our chatbot is able to generate responses by referring to proper entities from the collected knowledge. To validate the proposed approach, we build a movie conversation corpus on which the proposed approach significantly outperforms other four knowledge-grounded models.    
\end{abstract}

\section{Introduction}
Different from task-oriented dialogue assistants, social chatbots are not necessarily to solve problems. Rather, they are designed to engage and company users by chit-chat conversations~\cite{xiaoice}. It is critical for these chatbots to be aware of conversation-related knowledge especially when discussing topics. For example, when talking about a film, it is natural to mention its director and actors. 

\begin{figure}[!t]
\centering 
\includegraphics[height=1.6in]{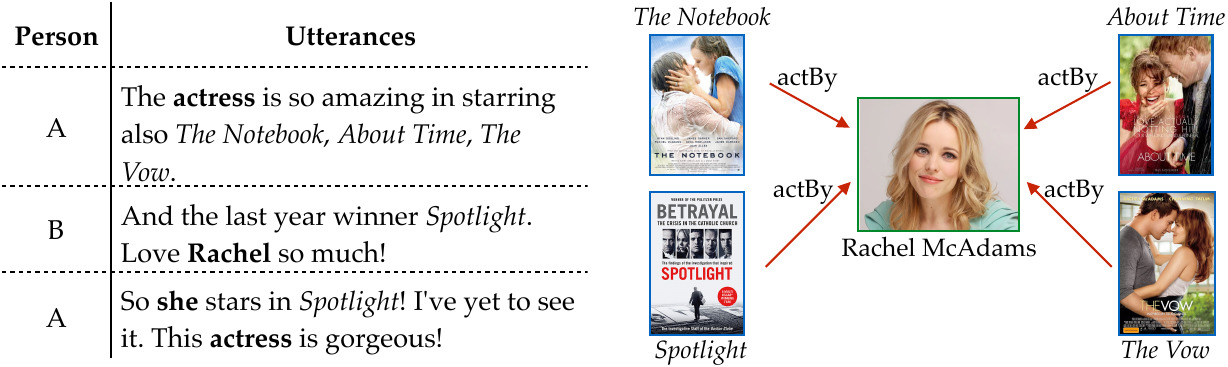}
\caption{A motivating conversation example on the topic: \emph{The Notebook}. Bold words indicate the underlying attribute, which is illustrated as the red arrows linking the films.}
\label{fig:example_dialog_english}
\end{figure}

Contextual knowledge is beneficial for context modeling. People often start and keep a conversation following a certain logic. See a real example in Figure~\ref{fig:example_dialog_english}. Given the film \emph{The Notebook} as the topic, user B imagines the new film \emph{Spotlight} because user A is talking about series of romance movies acted by the actress Rachel. As the \textbf{attribute} of the film, actress holds as the underlying logic link that naturally guarantees the coherence when moving forward the conversation. It is thus reasonable to equip chatbots with the ability to recognize the underlying attribute(s) for conversation understanding and link to related knowledge based on the recognized attribute(s). 

As another kind of contextual knowledge, \textbf{entities} are more important because they are extensively involved especially when people offer new information, provide supporting evidence, or refer to what has been mentioned. To facilitate response generation, chatbots need to also bear in mind related entities as candidates to be selected when responding to users. As revealed in the example, person A does not insist on the romance movies but moves on to the new one after person B introduces \emph{Spotlight}. In regard of the current context, the entities being considered in each turn may change. The larger the number of the candidate entities, the harder it will be for the chatbots to reason the most suitable one based on the current context. To ease this issue, our idea is to selectively collect the candidate entities using the recognized attributes to reduce the collection size. 

In this work, we explore how to collect and utilize contextual knowledge effectively for context modeling and response generation in chit-chat conversations. We develop a \textbf{M}ov\textbf{I}e \textbf{K}nowl\textbf{E}dge-grounded chatbot, namely \textsc{Mike}, equipped with a movie knowledge base (MKB). Given an input utterance(s) associate with a topic film, our \textsc{Mike} firstly recognizes the underlying attribute(s) and then collects candidate entities by starting from the mentioned entities and then propagating along the edge(s) of the recognized attribute(s). After equipped with necessary contextual knowledge, \textsc{Mike} captures the conversation context and generates responses based on an knowledge-enhanced encoder-decoder architecture~\cite{cho2014learning,sutskever2014sequence}. The encoder is enhanced with the detected attributes to compress the input utterance(s) into an attribute-aware context representation. 
The decoder is augmented with a pointer gate~\cite{vinyals2015pointer} to decide when to mention an entity and select from the candidates the most appropriate one based on the attribute-aware context. While previous work treat attributes and entities equally, our work is novel in discerning their differences and incorporating them in different manners. 
In this way, our chatbot \textsc{Mike} captures the conversation logic better with the help of contextual attributes, which in turn leads Mike to generate more coherent and entity-aware responses. 

To validate the effectiveness of the proposed approach, we build a new movie-related chit-chat conversation corpus, \textsc{Bili-Film}, collected from a large Chinese video platform. In brief, we highlight our contributions as follows:
\begin{itemize}
\item We identify external knowledge related to a conversation as contextual knowledge, and regard its necessities in both context representation and response generation.
\item We propose to utilize contextual attributes and entities in their own ways. The contextual attributes are contributing to capture the conversation logic for context modeling. The related entities are beneficial to generate responses when referring is needed.
\item We develop a novel movie knowledge-grounded chatbot, namely \textsc{Mike}, which firstly collects contextual knowledge from a MKB we build, and then generates entity-aware responses based on the attribute-aware context representation.
\item We build a movie conversation corpus, on which our \textsc{Mike} significantly outperforms other four knowledge-grounded models. The corpus will be released to the public. 
\end{itemize}

The rest of this paper is organized as follows. We discuss the related work in Section 2. Then, we describe our approach in detail in Section 3, which consists of contextual knowledge collector, attribute-aware context encoder, and entity-aware response decoder. Section 4 introduces the corpus we build and presents the experimental results on it, followed by further analysis in Section 5. Finally, we conclude in Section 6.

\section{Related Work}

Our work proposes a kind of knowledge-grounded conversation model, \textsc{Mike}, that is aimed at chatting with users based on related contextual knowledge collected from an associate KB. Our chatbot \textsc{Mike} is able to understand conversation context according to the detected attributes and generate entity-rich responses by properly referring to the related entities. In this section, we briefly review related work in this area. 

\subsection{Generation-based Conversation Models}
Due to the massive data and the development of neural networks, researchers have tried to build up chit-chat conversational systems using data-driven neural networks. Given a user utterance, the conversational systems are expected to return a proper response by either using retrieval techniques or generation techniques. To date, generation-based approaches have shown their effectivenesses. The pioneer work is~\cite{ritter2011data} that first formulates the response generation problem as Statistical Machine Translation (SMT), and reveals the feasibility of using massive Twitter data to build up a generation-based conversational model. 

From then, the majority of generation-based models apply the encoder-decoder architecture~\cite{sutskever2014sequence} which allows flexible modeling of user utterance and history utterances~\cite{Sordoni2015ANN,HRED,Tian2017HowContext}. Since history utterances often provide abundant information for conversation modeling, researchers have proposed extensive context-aware conversation models. The simplest way is to combine history utterances with the current one as the whole input using concatenation~\cite{lowe2015ubuntu,Sordoni2015ANN,Yan2016LearningTR,dailydialog}, pooling~\cite{Sordoni2015ANN}, or weighted combination~\cite{Tian2017HowContext}. More complicated way is to adopt hierarchical encoders by treating conversations as two-level sequences~\cite{HRED} which were extended with high-level latent variables to capture diversity in the conversation~\cite{VHRED,serban2017multiresolution,shen2017conditional,zhao2017discourse}. 

\subsection{Conditional Response Generation} 
Given a user input utterance, there often exists several proper kinds of responses. This is called the ``one-to-many'' problem in dialog response generation, and has been discussed in~\cite{li2016diversity,zhao2017discourse}. The diversity is resulted from a variety of influential factors. 

One typical factor is the conversation topic. It is very likely that people respond in conversation focusing on certain topics rather on rambling among unrelated issues. The idea of introducing topic information is to scope the response semantics during generation. Most existing work implement this idea by explicitly incorporating a topic variable estimated from the conversation data. 
\citet{xing2016topic} encodes both the input word embeddings and the topic keyword embeddings into a content encoder and a topic encoder, respectively. These two encoders then interact with each other in a joint attention mechanism to jointly determine the response decoding. However, it is hard to ensure that the topics learned from the external corpus are consistent with that in the conversation corpus. ~\citet{Yao2017TowardsIC} propose a content-introducing approach to generate a response based on a predicted keyword. \citet{Luan2016LSTMBC} comprises all previous dialog turns as a topic vector, which is then concatenated with hidden states together to predict the response tokens to be generated. 

\subsection{Knowledge-grounded Conversation Models}

In the line of combining conversational agents with knowledge bases, most work focus on developing task-oriented dialogue systems. These systems are associated with KBs on corresponding domains, e.g., restaurants, hotels, flights, etc. To generate responses, this kind of dialogue systems often detects user intentions at the beginning and then updates the dialogue states to fill related slots using the information from the corresponding KBs~\cite{wenN2N17,williams2016end,kbinfobot,LIDM}. 

Several efforts focus on improving user engagement by equipping non-task-oriented chatbots with KBs. The conversational model in~\cite{han2015exploiting} retrieves information KB to fill the response templates. \citet{yu2016strategy} handle possible breakdowns in chatbots by retrieving a short description to generate sentences. Some work attempt to incorporate implicit knowledge into chatbots to address the ``generic response'' problem. \citet{ghazvininejad2017knowledge} utilize external textual information as explicit knowledge for the chatbots. 
\citet{williams2017hybrid} present a model allowing developers
to express domain knowledge via software and action
templates. Most similar to our work is~\citet{zhu2017gends} that develops a dialogue system to talk about musics. Although grounding on knowledge, their system is focusing more on answering music-related questions. Moreover, each dialogue in their data is restricted to one singer. Differently, our approach is targeted at film-related chit-chats on various aspects rather than answering questions in movie domains.

\subsection{Knowledge Graph Representation}

As the fundamental pillars, knowledge bases and knowledge graphs (KBs and KGs) are emerging as important data sources for various applications. Typically, a knowledge graph (KG) is a multi-relational graph composed of entities (nodes) and relations (different types of edges). Each edge is represented as a triple of the form (head entity $e_h$, relation $r$, tail entity $e_t$). Such a triple is also called a fact, indicating that two entities are connected by a specific relation. For example, the triple \{The Notebook, {actBy}, Rachel McAdams\} describes the fact that the film \emph{The Notebook} is acted by Rachel McAdams. 

However, the symbolic nature of KGs impedes their applications. To tackle this issue, knowledge graph embedding models have been proposed to embed the relations and entities in a KG into low-dimensional continuous vector spaces. These KG embedding models can be roughly categorized into two groups: translation-based models and semantic matching models. Specifically, translation-based models learn the embeddings by calculating the plausibility of a fact as the distance between the two entities, usually after a translation carried out by the relation. Representative models are TransE~\cite{transe}, TransH~\cite{Wang2014KnowledgeGE}, TransR~\cite{Lin2015ModelingRP}. In TransE~\cite{transe}, the entity and relation embedding vectors are in the same space. In TransH~\cite{Wang2014KnowledgeGE}, entity embedding vectors are projected into a relation-specific hyperplane. In TransR~\cite{Lin2015ModelingRP}, entities are projected from the entity space to the relation space. \namecite{kgsurvey} summarizes other advanced knowledge embedding approaches. In this work, we embed our KG using the widely-adopted TransE model~\cite{transe}, and integrate the knowledge embeddings into conversation models in a novel way.

\section{Model}
\label{sec:mkb}
In this section, we describe the notation and framework of \textsc{Mike}. In two-party human-computer conversational systems, chatbots interact with users by returning proper responses. In particular, generation-based conversation models cast the problem of
response generation as a sequence to sequence
learning problem. Formally, conversation models take as input the combination of the current user utterance $\mathbf{u}^{T}$ and conversation histories $\{\mathbf{u}^1, \cdots, \mathbf{u}^{T-1}\}$, where $T$ is the turn number. Each utterance in the conversation is a sequence of words, a.k.a. $\mathbf{u^t} = \{x_1,\cdots,x_{N_t}\}$. Hence, chatbot is fed with a sequence of words $\mathbf{x} = \{x_1,\cdots,x_{N_x}\}$, and is required to generate a response $\mathbf{y} = \{y_1,\cdots,y_{N_y}\}$, where $N_x$ and $N_y$ are the token numbers.

\begin{figure*}[!t]
\centering 
\includegraphics[height=2.1in]{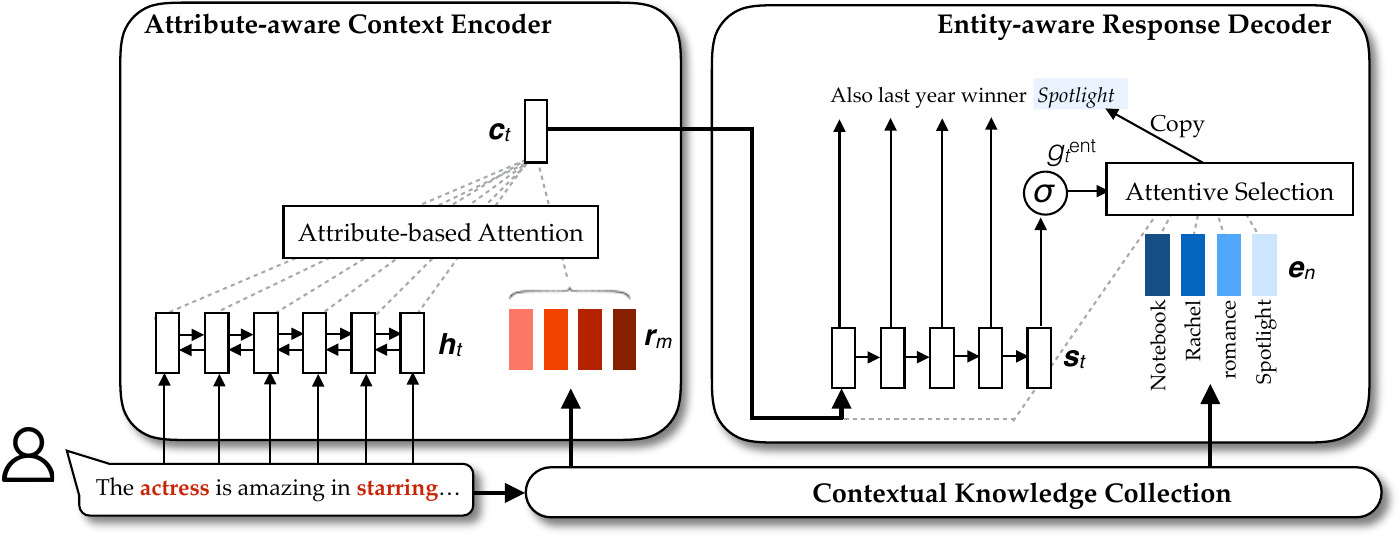}
\caption{The overview of the proposed chatbot \textsc{Mike}, which consists of three components. On the bottom is the prerequisite component, contextual knowledge collection, which will be illustrated detailedly in Figure~\ref{fig:knowledge}. The collected knowledge will be transfered into the encoder-decoder part, as shown in the top of the figure. Best viewed in color.}
\label{fig:model_detection}
\end{figure*}

\textsc{Mike} is a knowledge-grounded chatbot equipped with an associate knowledge base (KB) $\mathcal{K}$. Build upon the encoder-decoder architecture, \textsc{Mike} consists of three main components, as illustrated in Figure~\ref{fig:model_detection}:
\begin{itemize}
\item A contextual knowledge collector finds attributes and entities by linking the input sequence $\mathbf{x}$ to the associate KB $\mathcal{K}$. It detects the mentioned attributes from the input sequence, and collects entities relevant to the conversation. 
\item An attribute-aware encoder that transforms the input sequence of utterances $\mathbf{x}$ into a attribute-based representation by attending on the detected attributes.
\item An entity-aware decoder generates the final response by properly referring to the pre-collected entities.
\end{itemize}

With these three components, our approach firstly collects from $\mathcal{K}$ the contextual knowledge pertaining to the input $\mathbf{x}$, including the related attributes and entities. The detected attributes are used in the attribute-aware encoder to form an attribute-aware context representation, while the set of related entities are used as candidates to augment the entity-aware decoder. 

\subsection{Contextual Knowledge Collector} 
\begin{figure*}[!t]
\centering 
\includegraphics[height=3.7in]{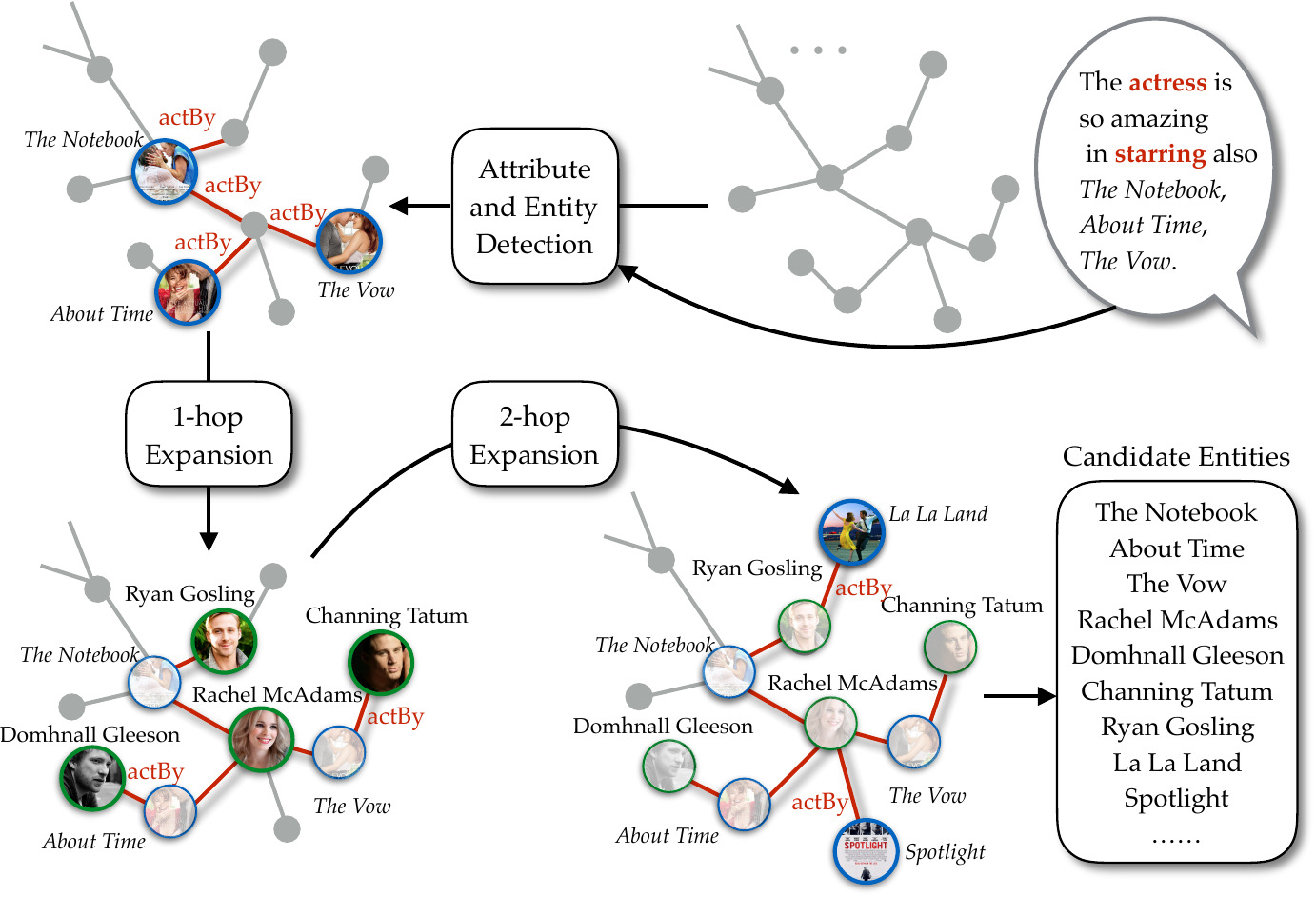}
\caption{Contextual Knowledge Collector.}
\label{fig:knowledge}
\end{figure*}

The prerequisite step is to collect the necessary knowledge from the associate MKB $\mathcal{K}$. In the left side of Figure~\ref{fig:knowledge}, a general sketch of $\mathcal{K}$ is depicted in the form of knowledge graph. The nodes are entities that are connected by the attributes on the edges. While there are often tens of thousands of facts stored in a KB, only a small fraction of them are related to a given conversation. It is more effective to scope a set of contextual knowledge by linking the input utterance(s) to the associate KB $\mathcal{K}$.\footnote{For efficiency, we retrieve a subgraph of the associate topic film, and perform knowledge discovery on the subgraph.} 

Given a conversation, the underlying logic is often indicated by the attribute information in the utterance(s). It is feasible to detect the attribute(s) $R_{\mathbf{x}}$ from the input utterance(s) using lexical patterns because they are often expressed regularly. For example, the words ``actress'' and ``starring'' (the red bold words in Figure~\ref{fig:knowledge}) indicate the attribute type \emph{actBy}. Similarly, we also detect a set of entities $E_{\mathbf{x}}$ mentioned in the input utterance(s) using entity linking techniques. By including the topic film into the set $E_{\mathbf{x}}$, we produce a set of seed entities $E_{\rm{seed}} = \{E_{\mathbf{x}}\cup e_{\rm{topic}}\}$, where $e_{\rm{topic}}$ is the topic entity.

However, it is insufficient to solely rely on the entities explicitly mentioned in the conversation. To expand the seed set $E_{\rm{seed}}$, we propose to collect more relevant entities $E_{r}$ by using the detected attribute(s) in $R_{\mathbf{x}}$. Concretely, we take each entity in $E_{\mathbf{x}}$ as head node $e_h$, and collect the entity on the tail node $e_t$ only if the relation $r_{e_h,e_t}$ between $e_h$ and $e_t$ matches with (one of) the detected attribute(s) $R_{\mathbf{x}}$. In this way, only the entities linked by the detected attribute(s) are collected to expand the entity set. We repeat this procedure by 2 times, which results in a 2-hop expansion as illustrated in Figure~\ref{fig:knowledge}. 

Notice that it is unreliable to expand the entity set using all the attributes in the KB, although it is straightforward to do so as in~\cite{zhu2017gends}. The larger the size of the entity set, the harder it will be for the chatbots to reason the most suitable when generating responses. Instead, we guide the entity set expansion based on the detected attributes, which is supposed to filter out noisy entities and eventually reduce the set size. The detected attributes will bias the entity expansion to collect those entities pertinent to the inherent conversation clue, and thus encourage more smooth and coherent conversations.

As a result, we collect the set of contextual knowledge related to a conversation including the set of detected attribute(s) $R=R_\mathbf{x}$ and the set of candidate entities {$E=\{E_{\mathbf{x}}\cup E_r\}$}. To fed this knowledge into the encoder-decoder conversational model, we encode the attributes and entities into dense representation. Specifically, we employ the knowledge graph embedding model TransE~\cite{transe} to transform them into dense embeddings, denoted as $\mathbf{r}_m$ and $\mathbf{e}_n$, respectively, where $\forall m \in \{1,\cdots,N_r\}$, $\forall n \in \{1,\cdots,N_e\}$. These attribute and entity embeddings are then fed to the encoder and decoder in their own ways, as shown in Figure~\ref{fig:model_detection}. The attribute embeddings are fed to the encoder to facilitate context modeling. The entity embeddings are served as candidates for the decoder to generate knowledgeable responses by selecting proper entities when referring is needed. 

\subsection{Attribute-aware Context Encoder} 
\label{sec:context}

Given the input sequence $\mathbf{x}$, we embed its tokens a Recurrent Neural Network, and then utilize the contextual attributes obtained in the first step to enhance the semantic representation.\footnote{When $\mathbf{x}$ consists of multiple turns of utterances, we concatenate them as a single, long utterance. We did not see clear empirical benefits using hierarchical context encoder.} This converts a sequence of inputs
$(\mathbf{x}_1, \mathbf{x}_2, \cdots, \mathbf{x}_t)$ to hidden states $(\mathbf{h}_1, \mathbf{h}_2, \cdots, \mathbf{h}_t)$. 

Typically, we adopt a bi-directional Gated Recurrent Unit (GRU)~\cite{gru} due to its advantage on learning long-term dependencies~\cite{bilstm}. The GRU cell consists of two gates, the update gate $\mathbf{z}_t$ and the reset gate $\mathbf{r}_t$, which are computed as follows:
\begin{align}
\mathbf{z}_t = \sigma(\mathbf{W}_z\mathbf{x}_t+\mathbf{U}_z\mathbf{h}_{t-1})\\
\mathbf{r}_t = \sigma(\mathbf{W}_r\mathbf{x}_t+\mathbf{U}_r\mathbf{h}_{t-1})
\end{align}

At each time step $t$, the update gate $\mathbf{z}_t$ controls how much the unit updates the content in the hidden states, whereas the reset gate $\mathbf{r}_t$ acts as a similar mechanism to allow the unit forget what has been previously computed. With these two gates, the hidden states at each time step $t$ is a linear interpolation computed as follows:
\begin{align}
\mathbf{h}_t &= (1-\mathbf{z}_t)\mathbf{h}_{t-1}+\mathbf{z}_t \mathbf{\tilde{h}}_t\label{eq:gru}\\
\mathbf{\tilde{h}}_t &= \textnormal{tanh}(\mathbf{W}_0\mathbf{x}+\mathbf{r})_t\odot(\mathbf{U}_0\mathbf{h}_{t-1}))
\end{align}
where $\odot$ is element-wise multiplication.

To encode the semantics from both the forward and backward of the input sequences, we adopt Bi-directional GRUs as our encoder basis. The Bi-directional GRUs are essentially a combination of two GRUs, one from the forward direction whereas the other from the backward. The resulting representation at each time step is the concatenation of each direction's state:
\begin{align}
\mathbf{h}_t = [\overleftarrow{\mathbf{h}_t},\overrightarrow{\mathbf{h}_t}]
\end{align}

Based on our preliminary studies, we propose to enrich the representation based on the detected attributes to form a context representation. As shown in the left bottom part of Figure~\ref{fig:model_detection}, we use an attribute-based attention mechanism~\cite{attention} to measure the semantic relevance between the utterance hidden states and the detected attributes. We compute the attribute-attention weights as:
\begin{align}
\alpha_{t} &\sim \textnormal{exp}(\mathbf{h}_t \mathbf{W}_1\bar{\mathbf{r}}) \label{eq:attention}\\
\notag \bar{\mathbf{r}} &= \frac{1}{N_r}\sum_{m=1}^{N_r} \mathbf{r}_m
\end{align}
where $\mathbf{W}_1$ is a learned matrix. Combined with the learned attention, the final context representation:
\begin{align}
\mathbf{c}_t = \alpha_{t}\mathbf{h}_t
\end{align}
which is then fed to the decoder. Intuitively, attribute-aware context encoder fuses the attribute information into the attribute-aware context representation, which is then used to initialize the hidden states of the decoder. When generating the responses, the attribute-aware context representation guides the decoder to prefer entities with similar representations, and thus allows the chatbot follow the underlying logic of the conversation.

\subsection{Entity-aware Response Decoder}

The last step is to properly respond by using the candidate entities related to the attributes. These candidate entities benefit the response generation when referring is needed. 

Basically, the
decoder is another GRU that takes as input the context representation $\mathbf{c}_t$ and the previously decoded token $y_{t-1}$ to update its hidden state $\mathbf{s}_t$ similar as Eq.~\ref{eq:gru}:
\begin{align}
\mathbf{s}_t = \textnormal{GRU}(\mathbf{s}_{t-1},[\mathbf{c}_t;y_{t-1}])
\end{align}
where $[;]$ is the concatenation operator of the two vectors. After obtaining the state vector at the current time step $t$, the decoder predicts the target word $y_t$ by performing a softmax classification based on its hidden state $\mathbf{s}_t$ and the context representation $\mathbf{c}_t$:
\begin{align*}
p^{\rm{gru}}(y_t | y_{1},\cdots,y_{t-1})&=f(y_{t-1},\mathbf{s}_{t-1}, \mathbf{c}_t)\\
&=\textnormal{softmax}(\mathbf{W}_o\mathbf{s}_t)
\end{align*}
where $\mathbf{W}_o$ is a parameter matrix. Hence, the decoder generates the response $\mathbf{y} = \{y_1, \cdots, y_{N_y}\}$ conditioned on the conversation context by maximizing the probability:
\begin{align}
p^{\rm{gru}}(y_1,\cdots,y_{N_y}|\mathbf{c}_t) &= p^{\rm{gru}}(y_1|\mathbf{c}_t)\prod_{t=2}^{N_y} p^{\rm{gru}}(y_t|y_1,\cdots,y_{t-1},\mathbf{c}_t)\\
&= p^{\rm{gru}}(y_1|\mathbf{c}_t)\prod_{t=2}^{N_y}
p(y_t|y_{t-1},\mathbf{s}_{t-1},\mathbf{c}_t)
\end{align}

To realize the entity-aware generation as illustrated in the bottom right of Figure~\ref{fig:model_detection}, we augment the decoder in the principle of pointer networks~\cite{vinyals2015pointer,Yang2016ReferenceAwareLM,copying}. Pointer networks have been demonstrated powerful on tackling out-of-vocabulary (OOV) words during generation. Previously, they are used to copy OOV words from the input sequences into the output sequences. Inspired by this idea, we adopt pointer networks to copy entities from external KB.

Concretely, we augment the decoder with a gating variable $g_t^{\textnormal{ent}}$ decides whether to generate an entity using $p^{\textnormal{ent}}$ or to omit a word from GRU language model using $p^{\textnormal{gru}}$ as follows:
\begin{align}
\notag p(y_t | y_{1},\cdots,y_{t-1})&=g_t^{\textnormal{ent}}p^{\textnormal{ent}}(y_t|y_{t-1},\mathbf{s}_t, \mathbf{c}_t,\mathbf{E})\\&+(1-g_t^{\textnormal{ent}})p^{\textnormal{gru}}(y_t|y_{t-1},\mathbf{s}_{t-1}, \mathbf{c}_t)
\end{align}
where $\mathbf{E}$ is the matrix stacking the candidate entity embeddings $\mathbf{e}_n$ obtained in the first step (Section 3.1). When the gate is ``open'', the decoder directly copies an entity by calculating the probability over the candidate entities $\mathbf{E}$. Otherwise, the decoder switches back to a vanilla GRU language model and omits a general word based on the softmax output. The gate $g_t^{\textnormal{ent}}$ is trained on the hidden state:
\begin{align}
g_t^{\textnormal{ent}} &= \sigma(\mathbf{W}_g\mathbf{s}_t)
\end{align}

The remaining is to learn which entity is to be selected by $p^{\textnormal{ent}}$ at each time step. We adopt another attention mechanism to approximate how close each entity is to the context, and obtain the attention weights $\boldsymbol{\beta}_t$ similar as Eq.~\ref{eq:attention}:
\begin{align}
\boldsymbol{\beta}_t \sim \text{exp}(\mathbf{{E}} \mathbf{W}_e \mathbf{c}_t)
\end{align}
Since the context representation $\mathbf{c}_t$ has been enriched by the attribute embeddings, the entities connecting with the detected attributes will have similar embeddings and then attract higher attention weights. The attended entities are naturally coherent to the conversation context. 

In a response, it is unlikely for an entity to be used twice even when multiple entities are mentioned together. It is thus necessary to be aware of what has already been generated when selecting the most appropriate entity at the moment. Inspired by the coverage attention~\cite{coverage}, we introduce a vector $\mathbf{a}_{tj} \in [0,1]$ that represents the probability each candidate entity already been mentioned in the response. The vector will be updated after a candidate entity is selected. In this way, the decoder has less chance to repeat an entity in the same response. 

Now the augmented decoder generates a candidate entity by: 
\begin{align}
p^{\textnormal{ent}}(y_t|y_{t-1}, \mathbf{s}_t, \mathbf{c}_t, \mathbf{E})=
\begin{cases}
\beta_{tj}, &\text{if } y_t =e_j\\
0, &\text{otherwise}
\end{cases}
\end{align}

The GRU language model $p^{\textnormal{gru}}$ is rather simple, and we adapt it to be aware of the film title by introducing the film title embedding in its content vector $\mathbf{c}_t$. This encourages the generation to stay focused.

\subsection{Training} 

After pre-collecting the candidate entities (by contextual knowledge collector), we are able to obtain supervision signals to train the switch gate $g^{\textnormal{ent}}_t$. We have:
\begin{align}
g^{\textnormal{ent}}_t =
\begin{cases}
1, &\text{if target word is a candidate entity}\\
0, &\text{otherwise}
\end{cases}
\end{align}

To train the model in the fully supervised manner, we have a training set of triples:
\begin{align*}
D = \{(X_1,Y_1,R_1,E_1)\}^{N_d}
\end{align*}
where $N_d$ is the number of training example, $X$ and $Y$ form the utterance-response pairs. Correspondingly, $R$ and $E$ are the sets of detected attributes and candidate entities, obtained using contextual knowledge collector in \textsc{Mike}.

Finally, we train model parameters by minimizing the negative log-likelihood objective as follows:
\begin{align}
NLL(D,\theta)= -\sum\sum_{i=1}^{N_d}\log p(Y_i|X_i,R_i,E_i)
\end{align}
The model parameters $\theta$ include the embeddings of
vocabulary, entities, relations, and the encoder-decoder components. Since the model is fully differential, we use stochastic gradient descent to back-propagate the gradients through the model components.

\section{Experiments}
In this section, we build a movie conversation corpus, on which we compare with 7 state-of-the-art conversational models to demonstrate the effectiveness of the proposed approach. As indicated by the automatic evaluations and human judgments, the proposed \textsc{Mike} outperforms other knowledge-grounded models significantly.

\subsection{Movie Conversation Corpus}
To validate the proposed approach, we build a novel movie conversation corpus, \textsc{Bili-Film}, which is collected from \textsc{Bilibili}, one of the largest Chinese video sharing and discussion platform.\footnote{\url{https://www.bilibili.com/v/cinephile/}} Although there are other movie discussion platforms,\footnote{i.e., \url{https://www.reddit.com/r/movies/}, \url{https://moviechat.org/}, \url{https://filmboards.com/}, etc.} the discussions on them are often focusing on detailed plots, and are too complex to learn. In contrast, the discussions \textsc{BiliBili} are more condense to capture.

On \textsc{BiliBili}, users can publish movie-related videos including the official trailers and films, as well as self-produced lens, montages, and narrations. Other users can discuss the videos by leaving new comments or responding to existing comments, as on typical forums. The comment threads between two users are the desired discussions we collect.

We define a seed set of 20 active publishers to crawl under their videos the discussions between two users. The corresponding film titles are also extracted from the video captions, and used as the discussion topics. We filter out some discussions that are meaningless or too long to learn. We maintain at most four speaker turns in all discussions. The statistics of our \textsc{Bili-Film} corpus is presented in Table~\ref{table:bili-film}.\footnote{The dataset will be released to the public.}

\begin{table}[!t]
\caption{Statistics of Corpus \textsc{Bili-film}.}
\begin{tabular}{r|r}
\hline
{Number of Total Conversations} & {8,368}\\
{Average Conversation Turns} & {3.6} \\
{Average Tokens Per Turn} & {27.8}\\\hline
{Number of Covered Films} & {162} \\
{Number of Covered Film Stars} & {239} \\
{Average Entities Per Turn} & {2.4} \\
{Unique Entities Per Conversation} & {3.1} \\\hline
\end{tabular}
\label{table:bili-film}
\end{table}

\subsection{Knowledge Base Construction}
To build a KB from scratch requires tedious effort. Instead, we build $\mathcal{K}$ based on \textsc{zhishi.me}~\cite{zhishi}, the largest Chinese knowledge base comprising comprehensive knowledge from three Chinese encyclopedias: Wikipedia Chinese version, Baidu Baike, and Hudong Baike.\footnote{\url{https://baike.baidu.com}, \url{https://www.hudong.com}} Despite being a general KB, \textsc{zhishi.me} has the largest coverage in movie domain compared to others like CN-DBPedia~\cite{cndbpedia}. 

\begin{figure*}[!t]
\includegraphics[height=1.4in]{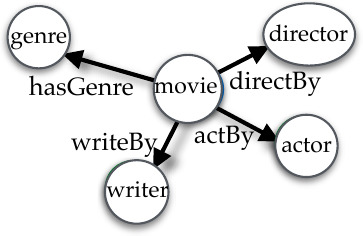}
\caption{The Schema of our MKB.}
\label{fig:schema}
\end{figure*}

We demonstrate the schema of our movie KB (MKB) $\mathcal{K}$ in Figure~\ref{fig:schema}. To acquire it, we extract from \textsc{zhishi.me} the triples containing either the attribute type \emph{actBy} or \emph{directBy}. This assumes to acquire all the films in it. As common practice, we add inverse attributes (i.e., \emph{actBy}$^{-1}$), and re-collect triples about these films according to five attribute types: \emph{hasAlias}, \emph{directBy}, \emph{actBy}, \emph{writeBy},\footnote{Some films are adapted from books, for example, the series of Harry Potter. In this case, we consider the writer of the book version.}, and \emph{hasGenre}. Correspondingly, there are five types of entities in our KB, i.e., film, director, actor(actress), writer, and genre.\footnote{These entities either refer to real-world things, e.g., \emph{The Notebook}, or to concept categories, e.g., romance movie. In this work, we treat them undistinguished.} 

As mentioned before, entity alias mining is crucial in our scenario. To improve the performance of entity discovery, we refine our MKB by extracting more alias information from an extra source. Although entities in \textsc{zhishi.me} already contain the attribute \textit{hasAlias}, they are sometimes out-of-date. To cover more, we also acquire alias from Douban Movie. For example, the famous Chinese director Stephen Chow (\begin{CJK*}{UTF8}{gkai}周星驰\end{CJK*}) are mostly mentioned with his nicknames \begin{CJK*}{UTF8}{gkai}周星星\end{CJK*} and \begin{CJK*}{UTF8}{gkai}星爷\end{CJK*}. However, the former one is missing in \textsc{zhishi.me} but found in Douban.\footnote{\url{https://movie.douban.com/celebrity/1048026/}} These additional nicknames are appended to facilitate entity linking.

\subsection{Compared Models}

To validate the effectiveness of \textsc{Mike} on incorporating knowledge into conversation modeling, we compare with the following models:

\begin{itemize}
\item\textsc{Attn-Enc-Dec}~\cite{attention}: It is the vanilla encoder-decoder approach with an attention mechanism, which is a widely adopted baseline. Both the encoder and decoder are vanilla GRUs~\cite{cho2014learning}. In this model, neither history utterances, nor extra knowledge is incorporated. This bare-bones model acts as a baseline to show the performance of encoder-decoder conversational systems without knowledge.

\item\textsc{Concat-Enc-Dec}~\cite{Sordoni2015ANN}: It is a extension of \textsc{Attn-Enc-Dec} where history utterances are concatenated along with the current input, and still without background knowledge.  

\item\textsc{HRED}~\cite{HRED}: This model considers conversation history in a hierarchical way, where a high-level RNN is built upon utterance-level RNN to capture conversation states.

\item\textsc{Fact-Enc-Dec}~\cite{ghazvininejad2017knowledge}: It is a knowledge-grounded conversation model that consumes relative textual facts as additional knowledge information. To fit it into our scenario, we use the films' one-sentence descriptions as the textual facts. By comparing with it, we aim at distinguishing the effects between the unstructured and structured knowledge.

\item\textsc{KB-Lstm}~\cite{knowledge_LSTM}: It identifies the knowledge related to the conversation and encodes the knowledge into conversation representation, which is similar with our idea. Differently, \textsc{KB-Lstm} only encodes the entities explicitly mentioned in the input utterance, and incorporates the entity encodings using concatenation operation in the encoder. On the contrary, we feed the context-relevant entities to the decoder for reasoning in response generation while our encoder takes the attribute information into account.

\item\textsc{KB-Lstm+}: We improve the above \textsc{KB-Lstm} model by incorporating also attributes information into the corresponding encoder. This is assumed to inject more knowledge implicitly and thus expand its knowledge scope. We denote this enhanced version as \textsc{KB-Lstm+}. 

\item\textsc{GenDS}~\cite{zhu2017gends}: It is the most similar approach to ours. \textsc{GenDS} shares a similar idea with ours that it ranks candidate entities collected from the retrieved facts to facilitate entity-aware response generation. Because candidates in \textsc{GenDS} also contain the entities implicitly mentioned in the input, it mainly differs with ours in how the candidate entities are selected. 

\end{itemize}

All models are implemented by TensorFlow~\cite{tensorflow}. To setup the experiments, the corpus is tokenized by Jieba Chinese word segmenter.\footnote{\url{https://github.com/fxsjy/jieba}} For TransE we utilize to encode our MKB, we adopt the implementation from KB2E~\cite{kb2e}.\footnote{\url{https://github.com/thunlp/KB2E}} We constrain the vocabulary to 25,000 words. The word embeddings are initialized with FastText vectors~\cite{fasttext}, and fine-tuned during training. The embeddings size are 300 and the hidden state vectors are 512. We set the mini-batch size as 32, and the learning rate to be 0.001 initially, which is decayed exponentially during training. We also clip gradients with norms larger than 0.5. Models are trained using Adam optimizer~\cite{adam}.

To evaluate our approach and the compared models, we adopt a variety of evaluation metrics that are widely used in previous works to examine the quality of the generated responses~\cite{sutskever2014sequence,lowe2015ubuntu,liu2016learning,shen2017conditional,xiaoice}, including both automatic evaluations and human judgments:
\begin{itemize}
\item BLEU-n: The BLEU scores indicating how much the generated responses is overlapped with the ground-truth response~\cite{bleu}; 
\item Dist-n: The distinct grams generated in the responses are indicative for the informativeness of the responses. The Dist-1 and Dist-2 scores, for unigrams and bigrams, are the ratios of types to tokens. This kind of diversity measurement is initially proposed by~\citet{li2016diversity} to examine the ``generic response problem'', which is then widely adopted in recent work~\cite{xing2016topic,wu2017neural}; 
\item Appropriateness and Grammar: According to previous studies~\cite{liu2016learning}, the aforementioned automatic metrics do not often correlate well with human judgments
in conversation generation tasks. To ease this issue, we also evaluate the models using human judgments. We first adopt two 3-scale human evaluation metrics, Appropriateness and Grammar, to judge the quality of the generated responses~\cite{shen2017conditional}; 
\item Precision and Recall: These two scores are used to examine the overlapping on knowledge-specific words, i.e., entity mentions in the generated responses~\cite{zhu2017gends}. The precision is the percentage of right generated entities in all generated entities, whereas the recall is of ground truth entities. These two metrics are calculated based on 100 manually annotated cases and are used to examine the ability of referring to the most relative entities. In this case, generating responses that contain irrelevant entities are not preferred. 
\end{itemize}

\subsection{Experimental Results}

\begin{table*}[!t]
\caption{Model Comparison Results. Bold numbers are the best performances.}
\begin{adjustbox}{max width=\textwidth} 
\begin{tabular}{r|cc|cc|cc|cc}
\toprule
& \multicolumn{4}{c|}{Automatic Evaluations}  & \multicolumn{4}{c}{Human Judgments}\\\hline
{\textbf {Model}}  & \textbf{BLEU-2} & \textbf{BLEU-3}  & \textbf{Dist-1} & \textbf{Dist-2} & \textbf{Appr.} & \textbf{Gram.} & \textbf{Prec.} & \textbf{Recall}\\\hline
\textsc{Attn-Enc-Dec} & 0.73&0.18&0.03&0.08&1.55&1.79&0.14&0.14\\
\textsc{Concat-Enc-Dec} & 0.76&0.20&0.03&0.10&1.58&1.72&0.14&0.15\\
\textsc{HRED} & 0.68&0.17&0.02&0.11&1.72&1.34&0.13&0.14\\\hline
\textsc{Fact-Enc-Dec}& 0.82&0.19&0.06&0.16&1.75&1.80&0.13&0.08\\
\textsc{KB-Lstm}&1.13&0.32&0.11&0.19&1.82&1.86&0.31&0.23\\
\textsc{KB-Lstm+}& 1.13&0.37&0.14&0.25&1.82&2.00&0.32&0.31\\\hline
\textsc{GenDS} & 1.09&0.68&0.16&\textbf{0.43}&1.96&2.03&0.37&0.38\\
\textsc{Mike}& \textbf{1.27}&\textbf{0.84}&\textbf{0.19}&0.40&\textbf{2.40}&\textbf{2.15}&\textbf{0.47}&\textbf{0.53}\\\bottomrule
\end{tabular}
\end{adjustbox}
\label{exp:compare}
\end{table*}

We first examine the significance of knowledge in chit-chat response generation. As shown in Table~\ref{exp:compare}, the three models in the first block (the first three rows) perform the worst. It is not surprising because they are the models that have no access to contextual knowledge. Their disappointing performances empirically support our research motivation that, it is necessary to incorporate background knowledge into chit-chat conversation models.

Since all other five models are equipped with background knowledge, we then compare them to find which mechanism utilize(s) the knowledge more effectively. According to the form of knowledge they consume, these five models can be further categorized into two groups: unstructural knowledge v.s. structural knowledge. It is obvious that \textsc{Fact-Enc-Dec} lags far from other knowledge-grounded models. It performs almost similar as \textsc{Attn-Enc-Dec} even it consumes extra knowledge. Notice that the fact knowledge it utilizes is represented in the form of natural language sentence, i.e., \emph{Titanic stars Leonardo DiCaprio and Kate Winslet as...}. Such unstructured representation impedes existing encoder-decoder models to exploit useful information from it and results in negligible improvement over the ``non-knowledge-aware'' \textsc{Attn-Enc-Dec}. On the contrary, \textsc{KB-Lstm}, \textsc{KB-Lstm+}, \textsc{GenDS} and our \textsc{Mike} utilize structural knowledge, i.e., the attributes and entities. Their better performances suggest that it is more effective to inject structural knowledge in encoder-decoder-based approaches.

Among the group of models that utilize structural knowledge, \textsc{KB-Lstm}, \textsc{KB-Lstm+} lag far from \textsc{Mike}. While both \textsc{KB-Lstm+} and \textsc{KB-Lstm+} employ attribute and entity information, \textsc{KB-Lstm+} results in small improvement over the original \textsc{KB-Lstm}. The difference between \textsc{KB-Lstm+} and \textsc{Mike} lies largely in the distinct mechanisms they utilize the attribute and entity information. \textsc{KB-Lstm+} comprises the attribute and entity information into a single vector and passes it to the RNN hidden state, which might be too elusive to guide high-quality response generation. Differently, \textsc{Mike} makes use of attribute and entities in different manners. \textsc{Mike} attends on the detected attributes to fuse the attribute information into context representation, and mentions the proper candidate entity(s) whenever the pointer gate is activated. The comparison results demonstrate the superiority of \textsc{Mike} on incorporating this knowledge.

\begin{table*}[!t]
\caption{Responses generated by the compared models, where generated entities are underlined. We translate Chinese input-response pairs to English. Note that input utterances are shorten due to limit space.}
\begin{tabular}{l|l}
\toprule
\multicolumn{2}{c}{\cellcolor{gray!25}{Example 1}}\\\hline
\textbf{Topic}& \begin{CJK*}{UTF8}{gkai}恋恋笔记本\end{CJK*} (\emph{The Notebook}) \\\hline
\multirow{2}*{\textbf{Input}}& \begin{CJK*}{UTF8}{gkai}男主帅，难怪女主动心了\end{CJK*} \\
&(The leading actor is so handsome that attracting the actress.) \\\hline
\textsc{Attn-Enc-Dec}& \begin{CJK*}{UTF8}{gkai}你没看懂\end{CJK*} (You didn't understand.) \\
\textsc{Fact-Enc-Dec}& \begin{CJK*}{UTF8}{gkai}男主电影好\end{CJK*} (The actor's film is good.) \\
\textsc{KB-LSTM+}& \begin{CJK*}{UTF8}{gkai}我很喜欢这个电影\end{CJK*} (I like this film.) \\
\textsc{GenDS}& \begin{CJK*}{UTF8}{gkai}这是\underline{瑞恩高斯林}\end{CJK*} (It's \underline{Ryan Gosling}.) \\
\textsc{Mike}& \begin{CJK*}{UTF8}{gkai}男主还演了\underline{爱乐之城}\end{CJK*} (He also stars in \underline{\emph{La La Land}}). \\
\hline
\multicolumn{2}{c}{\cellcolor{gray!25}{Example 2}}\\\hline
\textbf{Topic}& \begin{CJK*}{UTF8}{gkai}美人鱼\end{CJK*} (\emph{Mermaid})\\\hline
\multirow{2}*{\textbf{Input}}& \begin{CJK*}{UTF8}{gkai}说到画面和特效表现力，这片子水平难道不高吗？\end{CJK*} \\
& (Speaking of art and special effects, isn't it good?) \\\hline
\textsc{Attn-Enc-Dec}&  \begin{CJK*}{UTF8}{gkai}别提特效\end{CJK*} (Don't mention it.) \\
\textsc{Fact-Enc-Dec}& \begin{CJK*}{UTF8}{gkai}画面还凑合\end{CJK*} (The art is so-so.)\\
\textsc{KB-Lstm+} & \begin{CJK*}{UTF8}{gkai}\underline{周星驰}好\end{CJK*} (\underline{Stephen Chow} is good.)\\ 
\textsc{GenDS}& \begin{CJK*}{UTF8}{gkai}\underline{周星驰}一张电影票\end{CJK*} (\underline{Stephen Chow} a ticket.)\\
\textsc{Mike} & \begin{CJK*}{UTF8}{gkai}特效做的\underline{西游降魔篇}更好\end{CJK*} \\
& (\underline{\emph{Journey to the West}} is better in terms of special effects.)\\\bottomrule
\end{tabular}
\label{table:case}
\end{table*}

Overall speaking, \textsc{Mike} and \textsc{GenDS} are the best and second best models. They are the only models with explicit mechanism to generate entities from selected candidates. This proves the necessity of such mechanism for the chatbots. Different from \textsc{Mike}, \textsc{GenDS} retrieves entities by string matching the fact triples in the KB with the entities explicitly mentioned in the conversation utterances. In such unfiltered way, their candidate set might include noisy entities that are too tangential to the conversation context. As a result, \textsc{GenDS} has larger possibilities of attending on wrong, peripheral entities, and then generates unintelligible responses. On the contrary, \textsc{Mike} accesses to new entities $E_r$ linked by the detected attributes. The detected attributes will bias the entity expansion to collect implicit but material entities that closely related to the conversation. This novel strategy enables \textsc{Mike} to expand the conversation scope, and meanwhile limits the candidate set in a reasonable range. Drawing on the highest scores achieved by \textsc{Mike}, it proves the effectiveness of the strategy.

In general, the proposed \textsc{Mike} significantly outperforms the compared models in terms of almost all metrics except {Dist-2}. Especially, the automatic Distinct-n scores and human evaluation scores (Appr., Gram., Prec., and Recall.) indicate that the responses generated by our \textsc{Mike} are more diverse, fluent, and appropriate to the conversation context. 

For better understanding, we show some responses generated by five representatives in Table~\ref{table:case}. Clearly, the responses generated by \textsc{Att-Enc-Dec} and \textsc{Fact-Enc-Dec} are generic as they do not contain any named entities. Although \textsc{Fact-Enc-Dec} has access to the textual knowledge, it still generates less informative responses. This again demonstrates its inability of utilizing unstructured knowledge into response generation. We examine deeply into the generated responses in Example 2. By comparing the last three rows in Example 2, we are able to perceive the plausibility of responses generated by \textsc{KB-Lstm+}, \textsc{GenDS} and \textsc{Mike}. The topic film of Example 2 is \emph{Mermaid}, which is directed by \emph{Stephan Chow}. Despite that all the three models mention entities in the responses, \textsc{KB-Lstm+} and \textsc{GenDS} generates an entity, i.e., \emph{Stephen Chow} that is incoherent to the current input utterance. In this case, only \textsc{Mike} successfully refers to an proper entity, i.e., \emph{Journey to the West}, which is another film also directed by \emph{Stephen Chow}. Meanwhile, the response from \textsc{Mike} follows the conversation logic ``special effects'', which is a characteristic of the attribute \emph{directBy}. We attribute this success to the attribute-aware context representation in \textsc{Mike}.

\section{Model Analysis and Case Study}
\label{sec:exp_analysis}
\textsc{Mike} consists of three modules, i.e., contextual knowledge collector, attribute-aware encoder and entity-aware decoder. To examine the performance and contribution of each module, we conduct ablation studies and error analysis. We randomly select 100 test cases, and manually annotate the attribute and entities in the input utterance.

\subsection{Attribute and Entity Detection}
Note that in our case, the underlying attributes are often expressed regularly. Most entities mention in the text are movie-related. More importantly, we only care about those attributes and entities related to a specific given film. Hence, we use simple matching algorithms to separately detect attributes and entities from text. 

Given an input, the attributes are detected automatically by lexical patterns. For example, the appearance of ``actress'', ``starring in'', ``has a role of'' indicate the attribute \emph{actBy}. Based on the identified attributes, entity mentions are detected through string matching. Although some APIs are able to extract entities from short text, we find they are unreliable since the recall of a 100 test example are less than 10\%. More advanced approaches as in~\citet{knowledge_LSTM} might be of help but we leave it as future work. To improve matching quality, we clean the punctuations in advance., i.e., guillemets (\begin{CJK*}{UTF8}{gkai}《》\end{CJK*}), interpuncts (\begin{CJK*}{UTF8}{gkai}·\end{CJK*}) and quotation marks (\begin{CJK*}{UTF8}{gkai}“”\end{CJK*}). As a result, \begin{CJK*}{UTF8}{gkai}莱昂纳多·迪卡普里奥\end{CJK*} (Leonardo DiCaprio)=\begin{CJK*}{UTF8}{gkai}莱昂纳多-迪卡普里奥=莱昂纳多迪卡普里奥\end{CJK*}. 
To accelerate, we also segment the entity names and match them in segment units. In this case, ``Leonardo'' will also be successfully matched to ``Leonardo DiCaprio''. After detection, our {bili-film} corpus cover 226 film stars. And in average, there are 2.3 entities mentioned in each utterance, and 3.1 unique entities in each discussion. 

\begin{table}[!t]
\caption{Error Analysis. We do not count in the new entities not shown in the input utterance.}\label{table:error-detection}
\begin{tabular}{r|r|r|r|r|r}
\toprule
\multicolumn{3}{c|}{\textbf{Attribute}} &\multicolumn{3}{|c}{\textbf{Entity}}\\\hline
\textbf{Correct} & {94}&76.4\% &\textbf{Correct} & {128}&64.7\%\\
\textbf{Missing} & {12}&9.8\%& \textbf{Missing} & {65}&32.8\%\\
\textbf{Wrong} & {17}&13.8\% &\textbf{Wrong} & {5}&2.5\%\\
\bottomrule
\end{tabular}
\end{table}

\begin{table*}[!t]
\caption{Case Study of Attribute and Entity Detection. We translate Chinese input utterances to English.}
\begin{adjustbox}{max width=\textwidth} 
\begin{tabular}{l|l|l|l}
\toprule
\textbf{Case} & \textbf{Input Utterances} & \textbf{Detected} &\textbf{Truth}\\\hline
\multirow{3}*{1}&\begin{CJK*}{UTF8}{gkai}{我觉得就是因为星爷不是\textbf{主演}}\end{CJK*} &\multirow{3}*{actBy} &\multirow{3}*{directBy}\\
&(I think it is because that Stephen Chow is&& \\
&not the leading \textbf{actor}.)&&\\\hline
\multirow{3}*{2}&\begin{CJK*}{UTF8}{gkai}电影的\textbf{演员}是砖瓦，而特效仅仅是房子的装饰\end{CJK*} &{actBy}&\multirow{3}*{directBy}\\
&(For a film, the \textbf{actors} are tiles, while special effects &\multirow{2}*{directBy}&\\
&are only decorations.)&&\\\hline
\multirow{3}*{3}&\begin{CJK*}{UTF8}{gkai}老李子那个拿杯酒嘴角微翘的笑容堪称影片灵魂\end{CJK*} &\multirow{3}*{None}&\multirow{3}*{Leonardo}\\
&(The shot that ``Old Leo'' holds the glass and smiles &&\\
&is the soul of the film.)&&\\\hline
\multirow{3}*{4}&\begin{CJK*}{UTF8}{gkai}这个电影让我想起了周星驰的回魂夜\end{CJK*} &\multirow{3}*{None}&\multirow{2}*{Stephen Chow}\\
&(The film reminds me of Stephen Chow's &&\\
& \emph{Out of the Dark}.)&&{Out of the Dark}\\\bottomrule
\end{tabular}
\end{adjustbox}
\label{table:case_2}
\end{table*}

The performance of contextual knowledge collection is reported in Table~\ref{table:error-detection}. Since there often exist multiple attributes and entities in each utterance, the total number of the annotated ones are more than 100. It is shown that our detection accuracies are 76.4\% and 64.7\%, which are comparable to the performances in similar settings~\cite{zhang2016domain}. 

Although our scenario is much simpler, pattern matching techniques still face challenges. We show some cases in Table~\ref{table:case_2}. As shown in Case \#1, simple pattern matching will fail when the sentence has negative terms. This indicates that semantic parsing is needed when complex sentence grammar like concessive clause exists. Sometimes, the indicator word (pattern) is misleading as in Case \#2. Another kind of failure is caused by entity detection. In Case \#3, the model fails to link the mention ``Old Leo'' to the entity \emph{Leonardo} because the associate KB does not cover the alias ``Old Leo''. Note that the last case is about the conversation on the film \emph{Leon: The Professional} directed by Luc Besson. However, the user mentions the director Stephen Chow, which is not covered in the subgraph of \emph{Leon}. Theoretically, it is applicable to link entities based on the whole graph, which we leave as future work. 

\begin{table}[!t]
\caption{Ablation Studies. The first row corresponds to our full model.}
\begin{tabular}{r|c|c|c|c}
\hline
\multicolumn{1}{c|}{\textbf{Model}}& \textbf{BLEU-3} & \textbf{Dist-1} & \textbf{Prec.} & \textbf{Recall} \\\hline
\multicolumn{1}{c|}{\textsc{Mike}} & 0.84&0.19&0.47&0.53\\
\textbf{~~~-2HE} & 0.70&0.13&0.31 & 0.38 \\
\textbf{~~~-AAE} & 0.55&0.11&0.24&0.24 \\
\textbf{~~~-EAD} & 0.19&0.04&0.16&0.08\\
\hline
\end{tabular}
\label{table:ablation}
\end{table}

\subsection{Ablation Studies}

We perform additional ablation studies to investigate how important the following parts in our approach are: (1) the ``2-hop-expansion'' (2HE) solution in candidate entity selection; (2) the attribute-aware encoder (AAE); (3) and the entity-aware decoder (EAD). Table~\ref{table:ablation} presents the experimental results. For comparison purpose, we list the performance scores achieved by our full model \textsc{Mike} in the first row.

After removing the 2HE trick, as shown in the second row, the precisions and recalls will drop to 0.31 and 0.38 respectively, which indicates that it is necessary to expand the conversation scope by enlarging the candidate entities. As shown in Figure~\ref{fig:knowledge}, to encourage more diverse and richer conversation content, we treat the detected entities as seeds and add their neighboring entities that are linked by the detected attributes within 2-hops. Since all detected and added entities serve as the entity candidates to be selected in response generation, the proposed 2HE trick is beneficial to encourage more diverse and enriched responses. 

After replacing the attribute-aware encoder with a vanilla RNN encoder, the performance scores also decreases. This suggests that attribute-aware encoder is also crucial to facilitate conversation understanding by using the contextual attribute information. Intuitively, the attribute-aware context encoder fuses the attribute information into the attribute-aware context representation, which allows the chatbot follow the underlying logic of the conversation when generating the responses. 

Our approach degrades to standard Enc-Dec when all the special designs are removed. The remarkable gap between the scores in the last two rows are strong evidence for the necessity of the entity-aware decoder. Essentially, the decoder in the proposed \textsc{Mike} is a RNN language model augmented with the pointer gate~\cite{vinyals2015pointer,Yang2016ReferenceAwareLM,copying}. In this way, it ranks candidate entities collected from the associate knowledge base and thus generates more engaging and informative responses.

\section{Conclusions and Future Work}

In this paper we investigate conversation modeling using external knowledge, and propose a knowledge-grounded conversational model called \textsc{Mike}. Building upon the encoder-decoder architecture, the proposed \textsc{Mike} consists of three main components: (1) a contextual knowledge collector that performs knowledge discovery and transfer to link the associate KB with the given conversation; (2) a novel attribute-aware context encoder that represents current and history utterances using the collected attribute information; (3) a powerful entity-aware response decoder that generates informative responses by properly referring to suitable entities. With these three components, the proposed \textsc{Mike} are able to comprehend conversation logic using the detected attributes and respond to users more engagingly using the candidate entities.

On the movie conversation corpus \textsc{Bili-film} we build, we empirically demonstrate the effectiveness of \textsc{Mike}. It significantly outperforms other 7 state-of-the-art conversation models through both automatic evaluations and human judgments. The generated responses by \textsc{Mike} are the most plausible among the compared ones. We further conduct error analysis and ablation studies, and investigate the importance of each component in our approach. The overall experimental results reveal that attribute and entity information play distinguished and indispensable roles in conversation modeling, which have been neglected in previous research.

In the future, we plan to exploit more sources of external knowledge (i.e., film descriptions, plot summaries, the hot news about film artists), and explore effective ways to integrating these heterogeneous knowledge. Another promising direction is to model emotion and stance to sustain more friendly discussions.

\section*{Acknowledgments}

This work was supported by Research Grants Council of Hong Kong (PolyU 152094/14E, 152036/17E), National Natural Science Foundation of China (61672445 and 61272291) and The Hong Kong Polytechnic University (GYBP6, 4-BCB5, B-Q46C).

\starttwocolumn
\bibliography{compling_style}

\begin{thebibliography}{49}
\expandafter\ifx\csname natexlab\endcsname\relax\def\natexlab#1{#1}\fi

\bibitem[{Abadi et~al.(2016)Abadi, Barham, Chen, Chen, Davis, Dean, Devin,
  Ghemawat, Irving, Isard et~al.}]{tensorflow}
Abadi, Mart{\'\i}n, Paul Barham, Jianmin Chen, Zhifeng Chen, Andy Davis,
  Jeffrey Dean, Matthieu Devin, Sanjay Ghemawat, Geoffrey Irving, Michael
  Isard, et~al. 2016.
\newblock Tensorflow: A system for large-scale machine learning.
\newblock In \emph{OSDI}, volume~16, pages 265--283.

\bibitem[{Bahdanau, Cho, and Bengio(2014)}]{attention}
Bahdanau, Dzmitry, Kyunghyun Cho, and Yoshua Bengio. 2014.
\newblock Neural machine translation by jointly learning to align and
  translate.
\newblock \emph{arXiv preprint arXiv:1409.0473}.

\bibitem[{Bojanowski et~al.(2016)Bojanowski, Grave, Joulin, and
  Mikolov}]{fasttext}
Bojanowski, Piotr, Edouard Grave, Armand Joulin, and Tomas Mikolov. 2016.
\newblock Enriching word vectors with subword information.
\newblock \emph{arXiv preprint arXiv:1607.04606}.

\bibitem[{Bordes et~al.(2013)Bordes, Usunier, Garcia-Duran, Weston, and
  Yakhnenko}]{transe}
Bordes, Antoine, Nicolas Usunier, Alberto Garcia-Duran, Jason Weston, and
  Oksana Yakhnenko. 2013.
\newblock Translating embeddings for modeling multi-relational data.
\newblock In \emph{Advances in neural information processing systems}, pages
  2787--2795.

\bibitem[{Cho et~al.(2014{\natexlab{a}})Cho, van Merri{\"e}nboer, Bahdanau, and
  Bengio}]{gru}
Cho, Kyunghyun, Bart van Merri{\"e}nboer, Dzmitry Bahdanau, and Yoshua Bengio.
  2014{\natexlab{a}}.
\newblock On the properties of neural machine translation: Encoder--decoder
  approaches.
\newblock \emph{Syntax, Semantics and Structure in Statistical Translation},
  page 103.

\bibitem[{Cho et~al.(2014{\natexlab{b}})Cho, van Merrienboer, Gulcehre,
  Bahdanau, Bougares, Schwenk, and Bengio}]{cho2014learning}
Cho, Kyunghyun, Bart van Merrienboer, Caglar Gulcehre, Dzmitry Bahdanau, Fethi
  Bougares, Holger Schwenk, and Yoshua Bengio. 2014{\natexlab{b}}.
\newblock Learning phrase representations using rnn encoder--decoder for
  statistical machine translation.
\newblock In \emph{Proceedings of the 2014 Conference on Empirical Methods in
  Natural Language Processing (EMNLP)}, pages 1724--1734.

\bibitem[{Dhingra et~al.(2017)Dhingra, Li, Li, Gao, Chen, Ahmed, and
  Deng}]{kbinfobot}
Dhingra, Bhuwan, Lihong Li, Xiujun Li, Jianfeng Gao, Yun-Nung Chen, Faisal
  Ahmed, and Li~Deng. 2017.
\newblock Towards end-to-end reinforcement learning of dialogue agents for
  information access.
\newblock In \emph{Proceddings of ACL}.

\bibitem[{Ghazvininejad et~al.(2017)Ghazvininejad, Brockett, Chang, Dolan, Gao,
  Yih, and Galley}]{ghazvininejad2017knowledge}
Ghazvininejad, Marjan, Chris Brockett, Ming-Wei Chang, Bill Dolan, Jianfeng
  Gao, Wen-tau Yih, and Michel Galley. 2017.
\newblock A knowledge-grounded neural conversation model.
\newblock \emph{arXiv preprint arXiv:1702.01932}.

\bibitem[{Graves, Fern{\'a}ndez, and Schmidhuber(2005)}]{bilstm}
Graves, Alex, Santiago Fern{\'a}ndez, and J{\"u}rgen Schmidhuber. 2005.
\newblock Bidirectional lstm networks for improved phoneme classification and
  recognition.
\newblock In \emph{International Conference on Artificial Neural Networks},
  pages 799--804, Springer.

\bibitem[{Gu et~al.(2016)Gu, Lu, Li, and Li}]{copying}
Gu, Jiatao, Zhengdong Lu, Hang Li, and Victor~OK Li. 2016.
\newblock Incorporating copying mechanism in sequence-to-sequence learning.
\newblock In \emph{ACL}.

\bibitem[{Han et~al.(2015)Han, Bang, Ryu, and Lee}]{han2015exploiting}
Han, Sangdo, Jeesoo Bang, Seonghan Ryu, and Gary~Geunbae Lee. 2015.
\newblock Exploiting knowledge base to generate responses for natural language
  dialog listening agents.
\newblock In \emph{Proceedings of the 16th Annual Meeting of the Special
  Interest Group on Discourse and Dialogue}, pages 129--133.

\bibitem[{Kingma and Ba(2014)}]{adam}
Kingma, Diederik and Jimmy Ba. 2014.
\newblock Adam: A method for stochastic optimization.
\newblock \emph{International Conference for Learning Representations}.

\bibitem[{Li et~al.(2016)Li, Galley, Brockett, Gao, and
  Dolan}]{li2016diversity}
Li, Jiwei, Michel Galley, Chris Brockett, Jianfeng Gao, and Bill Dolan. 2016.
\newblock A diversity-promoting objective function for neural conversation
  models.
\newblock In \emph{Proceedings of the 2016 Conference of the North American
  Chapter of the Association for Computational Linguistics: Human Language
  Technologies}, pages 110--119.

\bibitem[{Li et~al.(2017)Li, Su, Shen, and Li}]{dailydialog}
Li, Yanran, Hui Su, Xiaoyu Shen, and Wenjie Li. 2017.
\newblock Dailydialog: A manually labelled multi-turn dialogue dataset.
\newblock In \emph{IJCNLP}.

\bibitem[{Lin et~al.(2015{\natexlab{a}})Lin, Liu, Luan, Sun, Rao, and
  Liu}]{Lin2015ModelingRP}
Lin, Yankai, Zhiyuan Liu, Huan-Bo Luan, Maosong Sun, Siwei Rao, and Song Liu.
  2015{\natexlab{a}}.
\newblock Modeling relation paths for representation learning of knowledge
  bases.
\newblock In \emph{EMNLP}.

\bibitem[{Lin et~al.(2015{\natexlab{b}})Lin, Liu, Sun, Liu, and Zhu}]{kb2e}
Lin, Yankai, Zhiyuan Liu, Maosong Sun, Yang Liu, and Xuan Zhu.
  2015{\natexlab{b}}.
\newblock Learning entity and relation embeddings for knowledge graph
  completion.
\newblock In \emph{AAAI}.

\bibitem[{Liu et~al.(2016)Liu, Sun, Lin, and Wang}]{liu2016learning}
Liu, Yang, Chengjie Sun, Lei Lin, and Xiaolong Wang. 2016.
\newblock Learning natural language inference using bidirectional lstm model
  and inner-attention.
\newblock \emph{arXiv preprint arXiv:1605.09090}.

\bibitem[{Lowe et~al.(2015)Lowe, Pow, Serban, and Pineau}]{lowe2015ubuntu}
Lowe, Ryan, Nissan Pow, Iulian~V Serban, and Joelle Pineau. 2015.
\newblock The ubuntu dialogue corpus: A large dataset for research in
  unstructured multi-turn dialogue systems.
\newblock In \emph{16th Annual Meeting of the Special Interest Group on
  Discourse and Dialogue}, page 285.

\bibitem[{Luan, Ji, and Ostendorf(2016)}]{Luan2016LSTMBC}
Luan, Yi, Yangfeng Ji, and Mari Ostendorf. 2016.
\newblock Lstm based conversation models.
\newblock \emph{CoRR}, abs/1603.09457.

\bibitem[{Niu et~al.(2011)Niu, Sun, Wang, Rong, Qi, and Yu}]{zhishi}
Niu, Xing, Xinruo Sun, Haofen Wang, Shu Rong, Guilin Qi, and Yong Yu. 2011.
\newblock Zhishi.me - weaving chinese linking open data.
\newblock In \emph{International Semantic Web Conference}.

\bibitem[{Papineni et~al.(2002)Papineni, Roukos, Ward, and Zhu}]{bleu}
Papineni, Kishore, Salim Roukos, Todd Ward, and Wei-Jing Zhu. 2002.
\newblock Bleu: a method for automatic evaluation of machine translation.
\newblock In \emph{Proceedings of ACL}, pages 311--318, Association for
  Computational Linguistics.

\bibitem[{Ritter, Cherry, and Dolan(2011)}]{ritter2011data}
Ritter, Alan, Colin Cherry, and William~B Dolan. 2011.
\newblock Data-driven response generation in social media.
\newblock In \emph{Proceedings of the conference on empirical methods in
  natural language processing}, pages 583--593, Association for Computational
  Linguistics.

\bibitem[{Serban et~al.(2016{\natexlab{a}})Serban, Sordoni, Bengio, Courville,
  and Pineau}]{HRED}
Serban, Iulian, Alessandro Sordoni, Yoshua Bengio, Aaron~C. Courville, and
  Joelle Pineau. 2016{\natexlab{a}}.
\newblock Building end-to-end dialogue systems using generative hierarchical
  neural network models.
\newblock In \emph{AAAI}.

\bibitem[{Serban et~al.(2017)Serban, Klinger, Tesauro, Talamadupula, Zhou,
  Bengio, and Courville}]{serban2017multiresolution}
Serban, Iulian~Vlad, Tim Klinger, Gerald Tesauro, Kartik Talamadupula, Bowen
  Zhou, Yoshua Bengio, and Aaron~C Courville. 2017.
\newblock Multiresolution recurrent neural networks: An application to dialogue
  response generation.
\newblock In \emph{AAAI}, pages 3288--3294.

\bibitem[{Serban et~al.(2016{\natexlab{b}})Serban, Sordoni, Lowe, Charlin,
  Pineau, Courville, and Bengio}]{VHRED}
Serban, Iulian~Vlad, Alessandro Sordoni, Ryan Lowe, Laurent Charlin, Joelle
  Pineau, Aaron Courville, and Yoshua Bengio. 2016{\natexlab{b}}.
\newblock A hierarchical latent variable encoder-decoder model for generating
  dialogues.
\newblock \emph{arXiv preprint arXiv:1605.06069}.

\bibitem[{Shen et~al.(2017)Shen, Su, Li, Li, Niu, Zhao, Aizawa, and
  Long}]{shen2017conditional}
Shen, Xiaoyu, Hui Su, Yanran Li, Wenjie Li, Shuzi Niu, Yang Zhao, Akiko Aizawa,
  and Guoping Long. 2017.
\newblock A conditional variational framework for dialog generation.
\newblock In \emph{Proceedings of the 55th Annual Meeting of the Association
  for Computational Linguistics (Volume 2: Short Papers)}, volume~2, pages
  504--509.

\bibitem[{Shum, He, and Li(2018)}]{xiaoice}
Shum, Heung-Yeung, Xiaodong He, and Di~Li. 2018.
\newblock From eliza to xiaoice: Challenges and opportunities with social
  chatbots.
\newblock \emph{arXiv preprint arXiv:1801.01957}.

\bibitem[{Sordoni et~al.(2015)Sordoni, Galley, Auli, Brockett, Ji, Mitchell,
  Nie, Gao, and Dolan}]{Sordoni2015ANN}
Sordoni, Alessandro, Michel Galley, Michael Auli, Chris Brockett, Yangfeng Ji,
  Margaret Mitchell, Jian-Yun Nie, Jianfeng Gao, and William~B. Dolan. 2015.
\newblock A neural network approach to context-sensitive generation of
  conversational responses.
\newblock In \emph{HLT-NAACL}.

\bibitem[{Sutskever, Vinyals, and Le(2014)}]{sutskever2014sequence}
Sutskever, Ilya, Oriol Vinyals, and Quoc~V Le. 2014.
\newblock Sequence to sequence learning with neural networks.
\newblock In \emph{Advances in neural information processing systems}, pages
  3104--3112.

\bibitem[{Tian et~al.(2017)Tian, Yan, Mou, Song, Feng, and
  Zhao}]{Tian2017HowContext}
Tian, Zhiliang, Rui Yan, Lili Mou, Yiping Song, Yansong Feng, and Dongyan Zhao.
  2017.
\newblock How to make context more useful? an empirical study on context-aware
  neural conversational models.
\newblock In \emph{Proceedings of the 55th Annual Meeting of the Association
  for Computational Linguistics (Volume 2: Short Papers)}, volume~2, pages
  231--236.

\bibitem[{Tu et~al.()Tu, Lu, Liu, Liu, and Li}]{coverage}
Tu, Zhaopeng, Zhengdong Lu, Yang Liu, Xiaohua Liu, and Hang Li.
\newblock Modeling coverage for neural machine translation.

\bibitem[{Vinyals, Fortunato, and Jaitly(2015)}]{vinyals2015pointer}
Vinyals, Oriol, Meire Fortunato, and Navdeep Jaitly. 2015.
\newblock Pointer networks.
\newblock In \emph{Advances in Neural Information Processing Systems}, pages
  2692--2700.

\bibitem[{Wang et~al.(2017)Wang, Mao, Wang, and Guo}]{kgsurvey}
Wang, Quan, Zhendong Mao, Bin Wang, and Li~Guo. 2017.
\newblock Knowledge graph embedding: A survey of approaches and applications.
\newblock \emph{IEEE Transactions on Knowledge and Data Engineering},
  29(12):2724--2743.

\bibitem[{Wang et~al.(2014)Wang, Zhang, Feng, and Chen}]{Wang2014KnowledgeGE}
Wang, Zhen, Jianwen Zhang, Jianlin Feng, and Zheng Chen. 2014.
\newblock Knowledge graph embedding by translating on hyperplanes.
\newblock In \emph{AAAI}.

\bibitem[{Wen et~al.(2017{\natexlab{a}})Wen, Miao, Blunsom, and Young}]{LIDM}
Wen, Tsung-Hsien, Yishu Miao, Phil Blunsom, and Steve Young.
  2017{\natexlab{a}}.
\newblock Latent intention dialogue models.
\newblock In \emph{ICML}, ICML'17, JMLR.org.

\bibitem[{Wen et~al.(2017{\natexlab{b}})Wen, Vandyke, Mrk\v{s}i\'{c}, Gasic,
  Rojas~Barahona, Su, Ultes, and Young}]{wenN2N17}
Wen, Tsung-Hsien, David Vandyke, Nikola Mrk\v{s}i\'{c}, Milica Gasic, Lina~M.
  Rojas~Barahona, Pei-Hao Su, Stefan Ultes, and Steve Young.
  2017{\natexlab{b}}.
\newblock A network-based end-to-end trainable task-oriented dialogue system.
\newblock In \emph{EACL}, pages 438--449, Association for Computational
  Linguistics, Valencia, Spain.

\bibitem[{Williams, Asadi, and Zweig(2017)}]{williams2017hybrid}
Williams, Jason~D, Kavosh Asadi, and Geoffrey Zweig. 2017.
\newblock Hybrid code networks: practical and efficient end-to-end dialog
  control with supervised and reinforcement learning.
\newblock \emph{arXiv preprint arXiv:1702.03274}.

\bibitem[{Williams and Zweig(2016)}]{williams2016end}
Williams, Jason~D and Geoffrey Zweig. 2016.
\newblock End-to-end lstm-based dialog control optimized with supervised and
  reinforcement learning.
\newblock \emph{arXiv preprint arXiv:1606.01269}.

\bibitem[{Wu et~al.(2017)Wu, Wu, Yang, Xu, Li, and Zhou}]{wu2017neural}
Wu, Yu, Wei Wu, Dejian Yang, Can Xu, Zhoujun Li, and Ming Zhou. 2017.
\newblock Neural response generation with dynamic vocabularies.
\newblock \emph{arXiv preprint arXiv:1711.11191}.

\bibitem[{Xing et~al.(2016)Xing, Wu, Wu, Liu, Huang, Zhou, and
  Ma}]{xing2016topic}
Xing, Chen, Wei Wu, Yu~Wu, Jie Liu, Yalou Huang, Ming Zhou, and Wei-Ying Ma.
  2016.
\newblock Topic augmented neural response generation with a joint attention
  mechanism.
\newblock \emph{URL http://arxiv. org/abs/1606.08340}.

\bibitem[{Xu et~al.(2017)Xu, Xu, Liang, Xie, Liang, Cui, and Xiao}]{cndbpedia}
Xu, Bo, Yong Xu, Jiaqing Liang, Chenhao Xie, Bin Liang, Wanyun Cui, and Yanghua
  Xiao. 2017.
\newblock Cn-dbpedia: A never-ending chinese knowledge extraction system.
\newblock In \emph{International Conference on Industrial, Engineering and
  Other Applications of Applied Intelligent Systems}, pages 428--438, Springer.

\bibitem[{Yan, Song, and Wu(2016)}]{Yan2016LearningTR}
Yan, Rui, Yiping Song, and Hua Wu. 2016.
\newblock Learning to respond with deep neural networks for retrieval-based
  human-computer conversation system.
\newblock In \emph{SIGIR}.

\bibitem[{Yang and Mitchell(2017)}]{knowledge_LSTM}
Yang, Bishan and Tom Mitchell. 2017.
\newblock Leveraging knowledge bases in lstms for improving machine reading.
\newblock In \emph{Proceedings of the 55th Annual Meeting of the Association
  for Computational Linguistics (Volume 1: Long Papers)}, volume~1, pages
  1436--1446.

\bibitem[{Yang et~al.(2016)Yang, Blunsom, Dyer, and
  Ling}]{Yang2016ReferenceAwareLM}
Yang, Zichao, Phil Blunsom, Chris Dyer, and Wang Ling. 2016.
\newblock Reference-aware language models.
\newblock In \emph{EMNLP}.

\bibitem[{Yao et~al.(2017)Yao, Zhang, Feng, Zhao, and Yan}]{Yao2017TowardsIC}
Yao, Lili, Yaoyuan Zhang, Yansong Feng, Dongyan Zhao, and Rui Yan. 2017.
\newblock Towards implicit content-introducing for generative short-text
  conversation systems.
\newblock In \emph{EMNLP}.

\bibitem[{Yu et~al.(2016)Yu, Xu, Black, and Rudnicky}]{yu2016strategy}
Yu, Zhou, Ziyu Xu, Alan~W Black, and Alexander Rudnicky. 2016.
\newblock Strategy and policy learning for non-task-oriented conversational
  systems.
\newblock In \emph{Proceedings of the 17th Annual Meeting of the Special
  Interest Group on Discourse and Dialogue}, pages 404--412.

\bibitem[{Zhang et~al.(2016)Zhang, Li, Li, Shi, Li, and Wang}]{zhang2016domain}
Zhang, Jiangtao, Juanzi Li, Xiao-Li Li, Yao Shi, Junpeng Li, and Zhigang Wang.
  2016.
\newblock Domain-specific entity linking via fake named entity detection.
\newblock In \emph{International Conference on Database Systems for Advanced
  Applications}, pages 101--116, Springer.

\bibitem[{Zhao, Zhao, and Eskenazi(2017)}]{zhao2017discourse}
Zhao, Tiancheng, Ran Zhao, and Maxine Eskenazi. 2017.
\newblock Learning discourse-level diversity for neural dialog models using
  conditional variational autoencoders.
\newblock In \emph{ACL}.

\bibitem[{Zhu et~al.(2017)Zhu, Mo, Zhang, Zhu, Peng, and Yang}]{zhu2017gends}
Zhu, Wenya, Kaixiang Mo, Yu~Zhang, Zhangbin Zhu, Xuezheng Peng, and Qiang Yang.
  2017.
\newblock Flexible end-to-end dialogue system for knowledge grounded
  conversation.
\newblock \emph{arXiv preprint arXiv:1709.04264}.

\end{thebibliography}

\end{document}